\documentclass[10pt,twocolumn,letterpaper]{article}

\usepackage[pagenumbers]{cvpr} 

\usepackage{subcaption} 

\usepackage{amsmath,amssymb,amsfonts}
\usepackage{algorithmic}
\usepackage{graphicx}
\usepackage{textcomp}
\usepackage{xcolor}
\usepackage{algorithm}
\usepackage{adjustbox}

\def\BibTeX{{\rm B\kern-.05em{\sc i\kern-.025em b}\kern-.08em
    T\kern-.1667em\lower.7ex\hbox{E}\kern-.125emX}}

\definecolor{cvprblue}{rgb}{0.21,0.49,0.74}
\usepackage[pagebackref,breaklinks,colorlinks,allcolors=cvprblue]{hyperref}

\def\paperID{*****} 
\def\confName{CVPR}
\def\confYear{2025}

\title{Token Sequence Compression for Efficient Multimodal Computing}

\author{Yasmine Omri\textsuperscript{*}\\
\and
Parth Shroff\textsuperscript{*}\\
\and
Thierry Tambe
}

\begin{document}
\twocolumn[{%
\maketitle
\begin{center}
\vspace*{-1.2cm}
    \textit{Stanford University} \\
    \{yomri, pmshroff, ttambe\}@stanford.edu \\
    {\footnotesize \textsuperscript{*} Authors with Equal Contribution}
\end{center}
\vspace{1em}
}]



\begin{abstract}
The exponential growth of Large Multimodal Models (LMMs) has driven advancements in cross-modal reasoning but at significant computational costs. In this work, we focus on visual language models. We highlight the redundancy and inefficiency in current vision encoders, and seek to construct an adaptive compression method for multimodal data. In this work, we characterize a panoply of visual token selection and merging approaches through both benchmarking and qualitative analysis. In particular, we demonstrate that simple cluster-level token aggregation outperforms prior state-of-the-art works in token selection and merging, including merging at the vision encoder level and attention-based approaches. We underline the redundancy in current vision encoders, and shed light on several puzzling trends regarding principles of visual token selection through cross-modal attention visualizations. This work is a first effort towards more effective encoding and processing of high-dimensional data, and paves the way for more scalable and sustainable multimodal systems.
\end{abstract}

\section{Introduction}
Large Multimodal Models (LMMs) have enabled processing and reasoning across various modalities, including images, video, and audio. LMMs excel in visual reasoning by integrating image, video, and text processing in VLMs \cite{liu2023improved, li2024llava, lin2024vila, li2023blip, adept2023fuyu, liu2024llava}. These models integrate visual information with textual semantics, allowing the model to perform tasks like visual question answering, image captioning, and multimodal inference.

However, the promising performance of VLMs largely relies on the large number of visual tokens generated during encoding. For instance, in LLaVA-1.5 \cite{liu2023improved}, the model generates 576 visual tokens, and in LLaVA-NeXT \cite{zhai2023siglip}, a 672x672 image can generate over 2880 visual tokens, while the text tokens are typically fewer (~10s per sentence). This disparity, combined with the quadratic scaling of attention operations with token count $O(T^2)$, results in significant computational overhead as the number of tokens increases.

The challenge becomes even more pronounced as we scale to incorporate additional modalities, such as video or audio, further increasing the token count and computational requirements \cite{ye2024xvila, wu2023nextgpt}.

\begin{figure}[t]
\hspace*{-3mm} 
\includegraphics[scale=0.6, trim=10 10 0 10, clip]{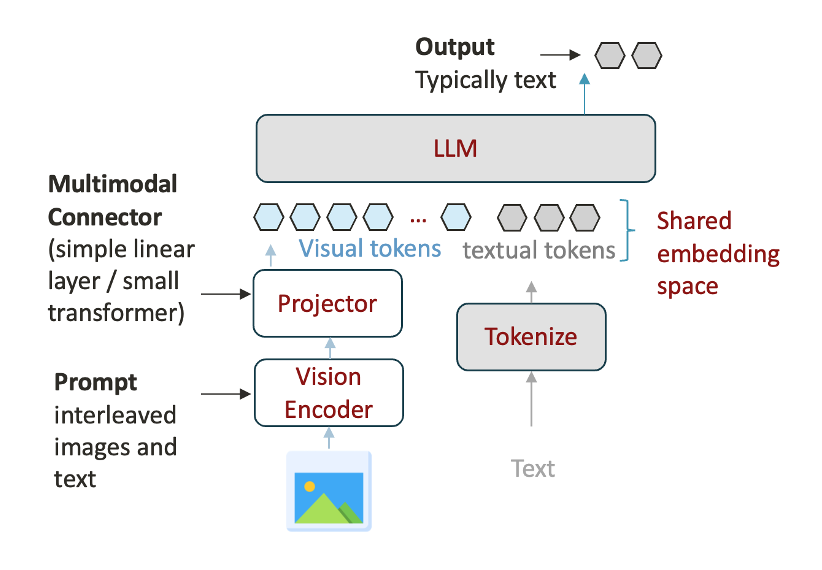}
\caption{Architecture of auto-regressive visual language models.}
\vspace{-0.5cm}
\label{vlm}
\end{figure}

This research focuses on addressing the challenge of hardware-efficient multimodal computing by investigating multimodal prompt compression. The central question of this paper is: \textit{How can we encode a multimodal prompt into a tokenized sequence that reduces the computational cost of multimodal inference, while maintaining satisfactory model accuracy?} In this paper, we examine existing and novel visual token selection methods. In particular, we consider both methods that assign importance scores to visual tokens for selection, and methods that do not. 
We underline surprising trends and inaccurate popular assumptions, and conclude with our ongoing and future directions in visual sequence compression.


\section{Background}
\begin{figure}[htbp]
\hspace*{-3mm} 
\includegraphics[scale=0.55]{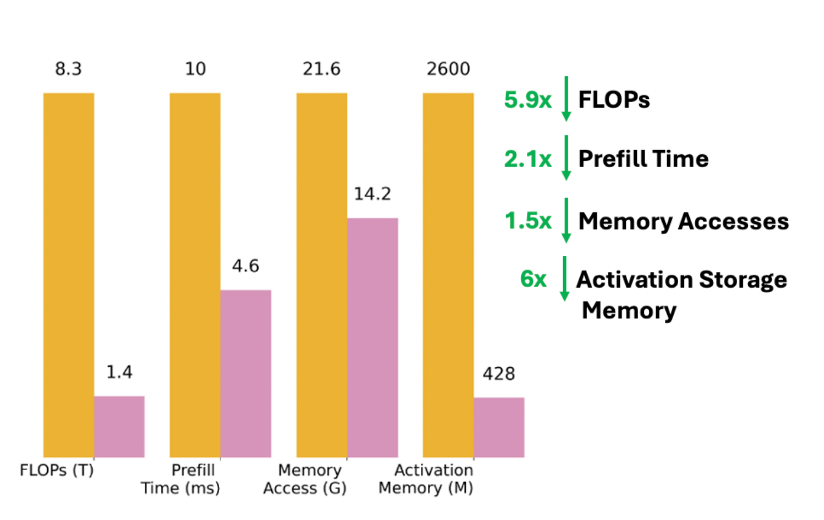}
\caption{Computational savings estimates at the LLM level from retaining 10\% of visual tokens, using LLMViewer\cite{yuan2024llm}.}
\label{compute}
\end{figure}

\textbf{Large Multimodal Model Architectures}\\ 
There are two main paradigms in VLMs. Cross-attention-based models, such as Flamingo and BLIP-2, use a dedicated component like Q-Former \cite{kim2024qformer} or Perceiver \cite{jaegle2021perceiver} to fuse the visual tokens into the language model’s attention mechanism. 
On the other hand, autoregressive models, such as Fuyu-8B, Palm-E, LLaVA, and VILA have gained popularity for their superior data efficiency and stronger alignment capabilities \cite{adept2023fuyu, liu2023improved, liu2024llava, li2024llava, lin2024vila}. Autoregressive VLMs typically consist of a vision encoder (e.g., CLIP \cite{radford2024clip} or SigLIP \cite{zhai2023siglip}), a projector to map visual tokens onto the shared space with text tokens, and a LLM that processes the integrated multimodal input, as illustrated in Figure~\ref{vlm}.\\ \\ 
\textbf{Computational Implications and Scalability} \\
Multimodal data dramatically increases the computational cost of language modeling, due to the inherently long context caused by the encoding of visual data and other complex modalities. This increases memory needs, prefill times, and quadratic self-attention costs. This directly impacts inference latency, energy consumption, and activation memory.
Fig. \ref{compute} demonstrates the dramatic effect of visual token count reduction on FLOPs, prefill time, memory access, and activation memory (numbers for NVIDIA H100 GPU using LLaMA2-7b based on the LLMViewer estimator from \cite{yuan2024llm}). Reducing the number of visual tokens while preserving critical information is essential for improving energy efficiency, enabling lower latency, and making VLMs more practical for real-world applications. Efficient token selection methods can mitigate these costs, ensuring that VLMs remain scalable without sacrificing accuracy.

\noindent\textbf{Related Work}\\
Some recent works have made advances in token reduction and merging for visual language models. Prumerge \cite{shang2024prumerge} and VisionZip \cite{yang2024visionzip} perform a text-agnostic visual token selection at the vision encoder level. These methods require fine-tuning at the whole LLM level and at the projector level respectively for best performance. ToMe \cite{tome} reduces the number of visual tokens inside the vision encoder by inserting selection and merging blocks between each attention block and MLP block, where merging is based on the cosine similarity between keys.
FastV \cite{chen2024fastv} and SparseVLM \cite{zhang2024sparsevlm} sparsify visual tokens as they progress through the network layers. These methods require the user to tune the token selection intensity through configuration knobs. FastV allows to prune away tokens for each LLM layer based on an attention-based ranking score. SparseVLM performs visual sparsification at each decoder layer, by considering their attention scores with important text tokens (excluding prepositions/pronouns). 



\section{Case Studies}
\begin{figure}[htbp]
\includegraphics[scale=0.3]{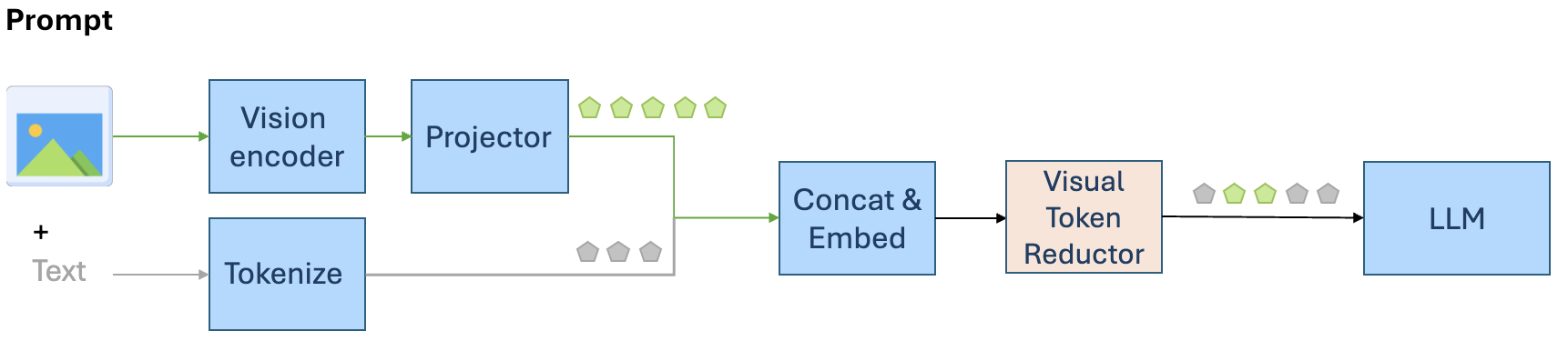}
\caption{Pipeline for dynamic training-free visual token sequence reduction pipeline.}
\label{pipeline}
\end{figure}

In this section, we investigate different algorithms for addressing the large visual token count and resulting computational burden. Unlike previous works, which select visual tokens inside the vision encoder or inside the LLM layers, we experiment with visual token sequence reduction algorithms that leverage the pre-LLM shared embedding space between text and image modalities as shown in Fig. \ref{pipeline}, and (1) experiment with text-image saliency formulations in Section \ref{tokensel}, (2) experiment with simple saliency-agnostic methods in Section \ref{random}. 

\subsection{Saliency-based Token Selection}\label{tokensel}
\begin{figure}[htbp]
\hspace*{-0.4cm} 
\includegraphics[width=1.05\columnwidth]{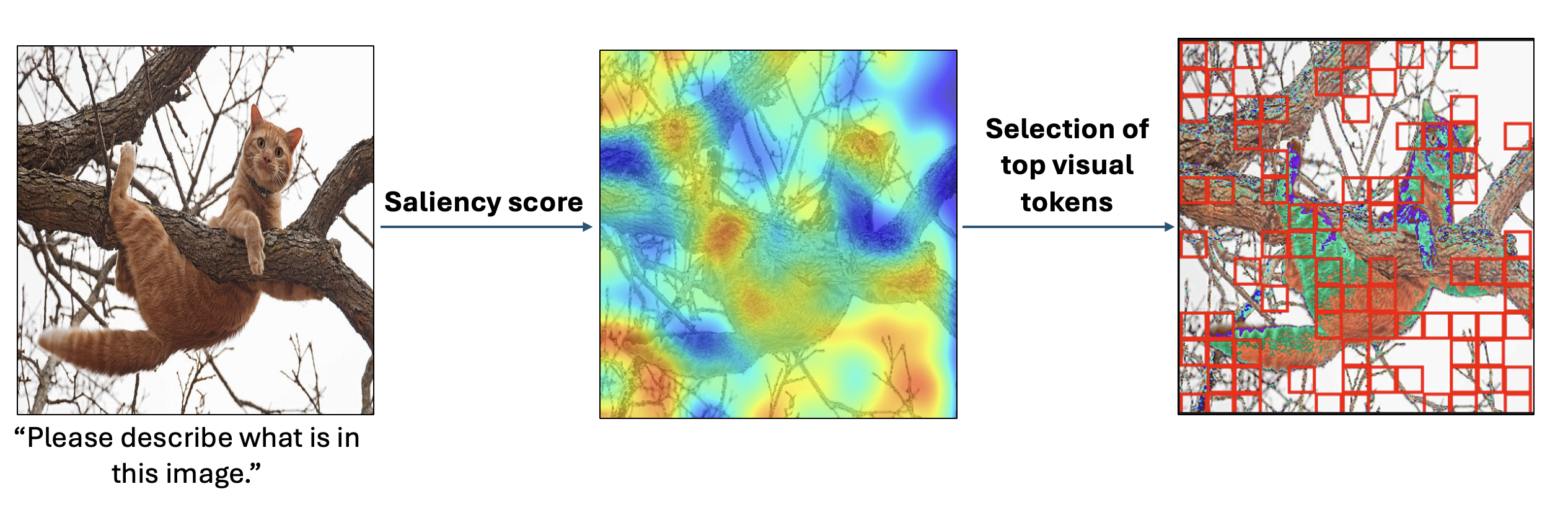}
\caption{Basic cross modality saliency visualization}
\label{fig:basiccms}
\end{figure}

In the first phase of our study, we investigate the correlation between cross-modal attention scores and semantic meaning. In particular, we seek to construct a visual token selection algorithm that uses the text prompt to inform which visual tokens are important and worth keeping. This measure of importance, or “saliency score” assigned to each visual token, is informed by the attention that each visual token attracts from the text tokens.
We hypothesize that such a saliency score can serve as a metric to inform a dynamic task-specific visual token selection algorithm. Various saliency score formulations were explored to manage the high dimensionality of the data and prevent oversmoothing. These included different approaches to handling attention heads, positional encoding, and augmentation techniques. 
We report results for two main variants:
\begin{itemize}
    \item \textbf{\textit{``Basic Saliency"}}, wherein visual tokens are ranked based on their importance, according to the saliency score formulation in Algorithm \ref{algo}. The top $x\%$ salient tokens from the original 576 visual tokens are retained and passed to the LLM along with the text tokens and system tokens. Refer to Fig. \ref{fig:basiccms} for a visualization of a basic saliency algorithm. 
    \item \textbf{\textit{``Cluster \& Saliency"}}, wherein visual tokens are first clustered via K-means++ based on their embeddings similarity. The top $x\%$ salient tokens from each cluster are retained.
\end{itemize}

\begin{algorithm}
\small
\scriptsize
\caption{Visual token saliency score formulation}
\label{algo}
\begin{algorithmic}
\REQUIRE Visual token embeddings, text token embeddings, query and weight matrices from the first layer

\STATE \textbf{Step 1: Compute saliency score for each visual token}
\FOR{each visual token $v_i$}
    \STATE Compute $k_{v_i} \cdot Q_{\text{text}}$ matrix
    \STATE Apply \textbf{softmax} over text dimension
    \STATE Obtain max over head dimension
    \STATE Compute average of these maxes for $v_i$ \COMMENT{Saliency score for $v_i$}
\ENDFOR
\STATE \textbf{Step 2: Retain the top visual tokens based on their saliency score rank}
\STATE \textbf{Step 3: Concatenate the retained visual tokens with the text tokens and system tokens. Pass this reduced token sequence to the LLM for processing.}

\end{algorithmic}
\end{algorithm}

There are two key intuitions for introducing clustering to saliency calculations: (1) it helps identify salient regions of the image by grouping coherent patterns, and (2) it denoises the representation by filtering out less informative content. We explore four clustering-based retention strategies that leverage these intutions which are enumerated in Table \ref{tab:clustering_variants} with relevant parameters and intuitions that they address. Once the clusters are constructed, each variant follows the following steps 

\begin{enumerate} 
    \item \textbf{Cluster Variant 1 – Static.} For each cluster, we compute a saliency score for every visual token using a predefined saliency function. Within each cluster, the top $x\%$ of tokens (based on saliency) are retained. This ensures that the final retained set comprises $x\%$ of the total visual tokens, distributed across all clusters. The key idea is that each cluster captures distinct regions of the image, and enforcing per-cluster retention encourages spatial diversity in the retained set.
    \item \textbf{Cluster Variant 2 – Dynamic.} Instead of a fixed per-cluster retention rate, this variant adapts retention based on the overall importance of each cluster. Specifically, the average saliency score is computed for each cluster, and a softmax function is applied to produce normalized weights $\{w_i\}$. These weights determine the relative importance of each cluster. The algorithm then retains the top $\lambda w_i\%$ of tokens from each cluster $c_i$, given some user-input scale $\lambda$. This formulation allows more informative clusters to retain a proportionally larger number of tokens.

    \item \textbf{Cluster Variant 3 – Coarse Aggregation with Saliency.} This variant aims to preserve global information by aggregating non-salient tokens rather than discarding them. After computing saliency scores, the top $x\%$ of tokens are retained from each cluster as in Variant 1. The remaining $(100 - x)\%$ of tokens are then averaged to form a single aggregated token per cluster. This aggregation is performed by computing both the mean of the token embeddings and their positional coordinates. The final output includes the $x\%$ retained tokens and $k$ aggregated tokens (one per cluster), resulting in a total of approximately $(x + \delta)\%$ retained tokens, where $\delta$ depends on the number of clusters $k$ and token count.

\end{enumerate}

Motivated by recent work such as PruMerge ~\cite{shang2024prumerge} demonstrating the semantic advantages of key embeddings, we also explored the clustering over keys. However, the results were slightly worse than all of the other variants and further analysis is required to fully understand the slightly degraded performance.

\begin{table*}[t]
\centering
\Large  
\caption{Comparison of clustering-based token retention variants.}
\label{tab:clustering_variants}
\begin{adjustbox}{width=\textwidth}
\begin{tabular}{@{}llccccp{5.5cm}@{}}
\toprule
\textbf{Number} & \textbf{Variant} & \textbf{Parameters} & \textbf{Clustering Basis} & \textbf{Saliency Strategy} & \textbf{Retention Method} & \textbf{Key Intuition} \\
\midrule
1 & Static & $k, x$ & Embedding similarity & Per-token within cluster & Top $x$\% per cluster for $k$ clusters & Clusters reflect distinct salient regions \\
2 & Dynamic & $\lambda$ & Embedding similarity & Average per-cluster & Top $\lambda w_i$\% per cluster $c_i$ (with $w_i$ via softmax) & Some clusters are more informative than others \\
3 & Coarse Aggregation with Saliency & $k, x$ & Embedding similarity & Per-token within cluster & Top $x$\% retained, remaining $(100{-}x)$\% aggregated per $k$ clusters & Retains useful information by aggregating less salient tokens instead of removing them. \\
\bottomrule
\end{tabular}
\end{adjustbox}
\end{table*}

\begin{figure}[t]
    \centering
    \begin{subfigure}[b]{\columnwidth}
        \centering
        \includegraphics[width=1.2\textwidth]{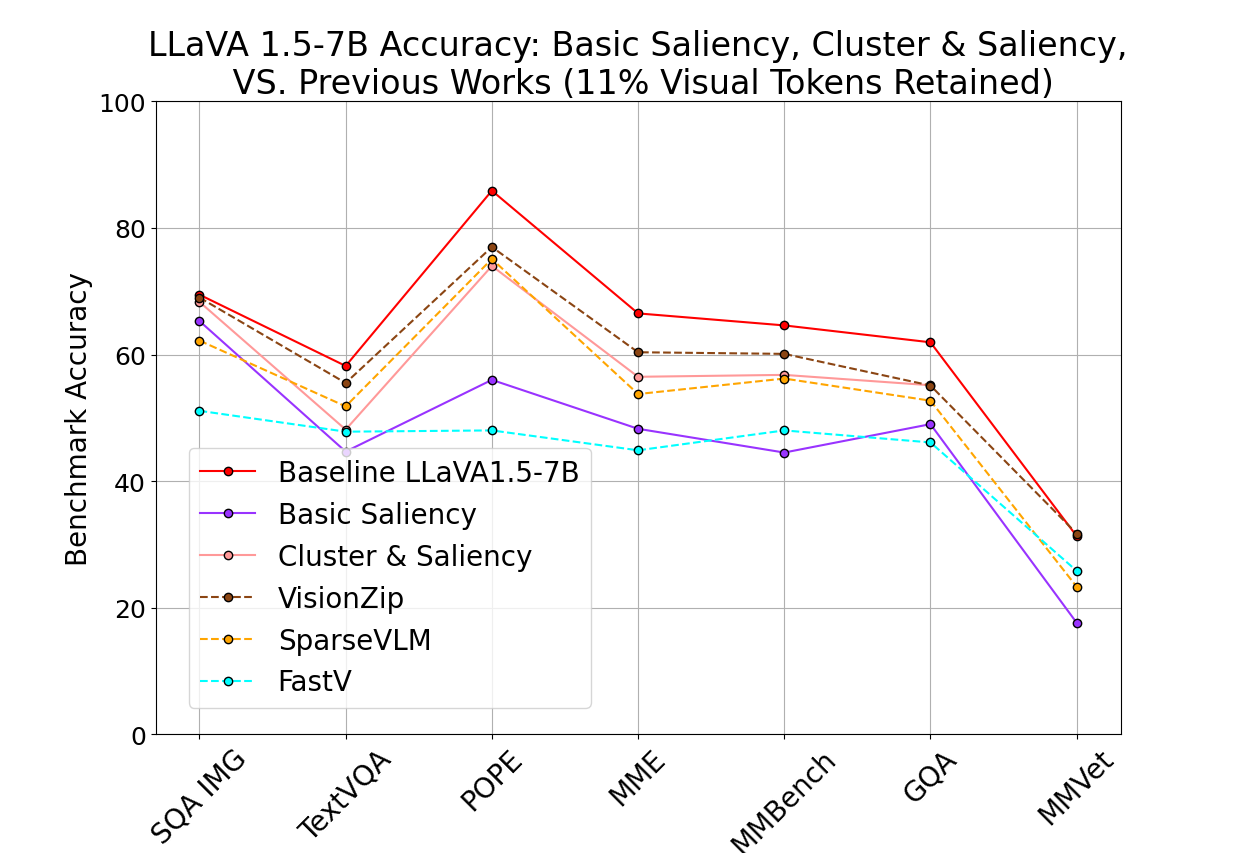}
        \caption{}
        \label{fig:performance_chart}
    \end{subfigure}
    



    \begin{subfigure}[b]{\columnwidth}
        \centering
        \includegraphics[width=1.1\textwidth]{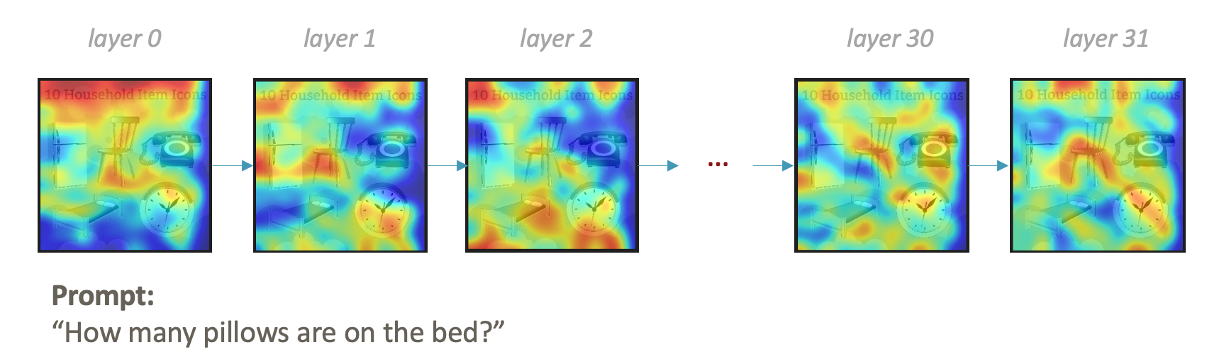}
        \caption{}
        \label{fig:layer_heatmaps}
    \end{subfigure}

    \caption{(a) Performance of LLaVA1.5-7B with 11\% of the salient visual tokens. (b) Layer-wise visualization of saliency heatmaps }
    \label{fig:full_analysis}
\end{figure}

We report accuracy results for LLaVA 1.5-7B on 7 visual language benchmarks \cite{lu2022sqa, singh2019textvqa, li2023objecthallucination, fu2023mme, yu2024mmvet, liu2023mmbench, hudson2019gqa} for both \textit{Basic Saliency} and \textit{Cluster \& Saliency} in Fig. \ref{fig:performance_chart} for a visual token retention percentage of 11\% (a popular retention percentage used by the previous SoTA works). Note that for \textit{Cluster \& Saliency}, variant 1 was used since one can statically fix the percentage retained with the input $x$ parameter.  
Despite its intuitiveness, the basic saliency metric performs poorly compared to most SoTA works such as SparseVLM and VisionZip. 
However, we find that enhancing the saliency metric with clustering dramatically improves the results. In particular, clustering helps denoise the image and increases the spatial diversity of selected tokens.

We also present a comparative analysis of the four clustering variants' performance in Table \ref{table_cluster_variant}. It should be noted that Variants 2 and 3 exhibit retention percentages that deviate slightly from the $11\%$ benchmark, as Variant 2 employs a dynamic clustering approach without fixed k-parameter inputs, while Variant 3 deliberately retains $(x + \delta)\%$ of tokens. Although parameter optimization could potentially align these variants more precisely with the $11\%$ target, our empirical evidence indicates no statistically significant performance differential among variants when controlling for retention percentage. Indeed, performance correlates strongly with the proportion of retained tokens—an intuitive finding suggesting that information preservation is the primary determinant of performance rather than clustering methodology. These results informed our subsequent investigation into the fundamental limitations of saliency-based approaches. While integrating clustering techniques with basic saliency methods yielded measurable performance improvements, the highest-performing algorithms identified in the following section did not incorporate cross-modality saliency metrics for compression, challenging our initial hypotheses.

\begin{table*}[t]
\centering
\scriptsize
\caption{Performance of cluster variants} 
\label{table_cluster_variant}
\resizebox{\textwidth}{!}{
\begin{tabular}{@{}llccccc|cccc@{}}
\toprule
\textbf{Method} & \textbf{Retained \%} & \textbf{SQA IMG} & \textbf{TextVQA} & \textbf{POPE} & \textbf{Vizwiz} & \textbf{MME} & \textbf{MMBench} & \textbf{MMVet} & \textbf{GQA} & \textbf{VQAV2} \\
\midrule
\multicolumn{11}{l}{\textbf{Baseline}} \\
LLaVA1.5-7B & 11 & 69.46 & 58.19 & 85.9 & 50.08 & 1862 & 64.6 & 31.3 & 61.93 & 78.52 \\
\midrule
\multicolumn{11}{l}{\textbf{Cluster Variants}} \\
Cluster Variant 1 ($k = 20$) & 11 & 67.58 & 48.61 & 72.83 & 53.16 & 1589.67 & 56.44 & 26.3 & 55 & 49.58 \\
Cluster Variant 2 ($\lambda = 1$) & $\approx$6 & 67.13 & 46.96 & 68.01 & 53.06 & 1498 & 52.83 & 23.9 & 53.11 & 20.36 \\
Cluster Variant 2 ($\lambda = 1.5$) & $\approx$9 & 67.43 & 47.57 & 70.49 & 53.52 & 1538.423 & 55 & 25.2 & 54.09 & 65.35 \\
Cluster Variant 3 ($k = 20$) & $\approx$13 & 68.77 & 49.54 & 75.47 & 54.05 & 1643 & 59.36 & 28.9 & 55.49 & 65.35 \\
Cluster Variant 3 ($k = 40$) & $\approx$15 & 69.46 & 52.97 & 81.28 & 53.99 & 1703.51 & 60.46 & 31.9 & 57.38 & 65.35 \\
\bottomrule
\end{tabular}
}
\end{table*}

\textbf{Puzzling Trends.}
To better understand the limitations of attention-based saliency scores, we visualize the saliency score heatmap for various combinations of images and prompts. The goal is to understand which regions of images are deemed important and retained based on different prompts. 
We study the saliency heatmaps for an extensive amount of prompt-image combinations, and observe the following two puzzling trends:
\\
\begin{itemize}
    \item (Trend 1) The salient regions most often do not coincide with the intuitively informative regions of the image. 
    Additionally, saliency rankings are volatile and do not converge across the model layers, as suggested by Fig. \ref{fig:layer_heatmaps}. This calls into question the viability of the growing reliance on attention-based importance metrics for visual token selection.
    \item (Trend 2) The saliency heatmap exhibits very little variance with different prompts: even though the goal of attention-based metrics is to use text to inform dynamic prompt-based visual token selection, the saliency ranking of visual tokens barely changes with the text prompt. This suggests that certain visual tokens inherently attract more attention based on their embeddings.  
\end{itemize}

\vspace{1cm}
\subsection{Random Sampling, Spatial Sampling, and Cluster-based Embedding Aggregation}\label{random}
In the second phase of our study, we investigate simple token selection methods that do not attribute importance to different visual tokens. We benchmark these approaches, and compare them to various prior works, including the saliency-based algorithms presented in Section \ref{tokensel}. 

In particular, we benchmark 3 different importance-agnostic selection methods:
\\
\begin{itemize}
    \item (1) \textbf{\textit{Random Sampling}} of visual tokens.
    \item (2) \textbf{\textit{Spatial Sampling}} of visual tokens (uniform stride).
    \item (3) \textbf{\textit{Cluster-based Embedding Aggregation: ``Cluster \& Aggregate"}}: After the vision encoder and the projector, the visual tokens are clustered using K-means++ based on their embedding similarity. Except, instead of using saliency to select some of the tokens from each cluster, we simply aggregate all the tokens from each cluster into one aggregate token per cluster, by averaging the token embedding vectors. The resulting aggregate embeddings are concatenated with a random order. An alternative approach is to concatenate the aggregate embeddings based on the average position of tokens within each cluster. However, random insertion is a simpler method and proved to yield comparable results.
   
\end{itemize}

\begin{figure}[htbp]
\includegraphics[width=1.1\columnwidth]{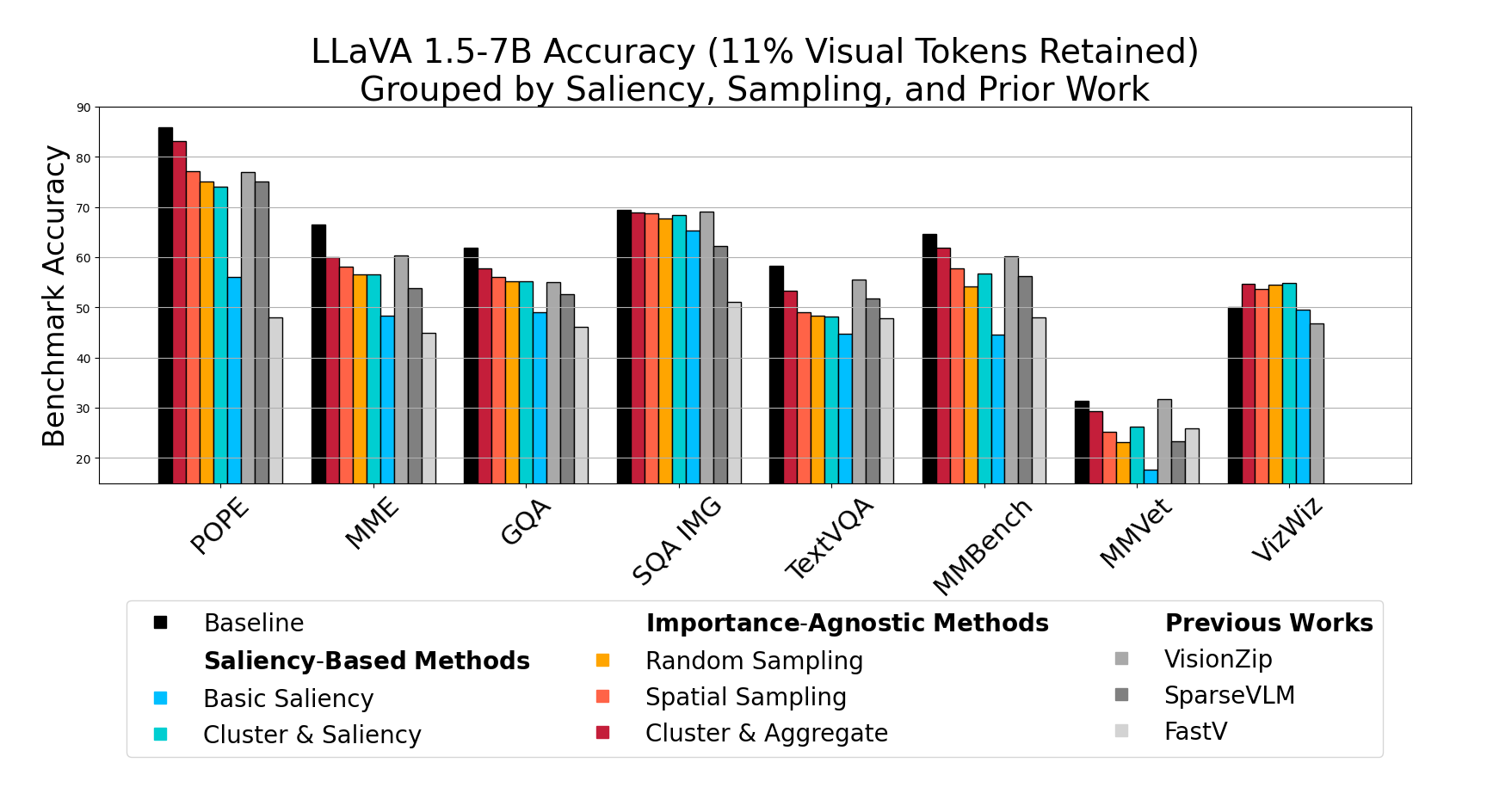}
\caption{Comparison of importance-agnostic selection methods and saliency-based selection methods based on the accuracy of LLaVA 1.5-7B (11\% of visual tokens retained)}
\label{ElevenPercentResults}
\end{figure}

As shown in Fig.~\ref{ElevenPercentResults}, the Cluster \& Aggregate method (Red) consistently outperforms all saliency-based variants (Blue), as well as prior finetuning-free state-of-the-art token selection methods (Gray). In particular, Cluster \& Aggregate achieves superior or comparable performance to the finetuning-free version of VisionZip, and outperforms it on average on the 8 benchmarks tested. Additionally, we observe that Random Sampling and, in particular, Spatial Sampling achieve competitive performance, surpassing most baselines despite their simplicity. Interestingly, we also notice that fewer visual tokens can actually result in higher accuracy. This is the case for VizWiz where the majority of token selection methods results in an improved visual question answering baseline.
\\
The surprisingly high performance of cluster-based embedding aggregation, spatial sampling, and random sampling when compared to importance-based sampling methods further call for revisiting the popular and growing intuition that visual attention scores (either with text tokens or with the CLS token) are correlated with token importance in downstream accuracy. This also underlines a high redundancy and deficiency in the encoding process. On one hand, some tokens inherently attract more attention despite their low semantic importance. On the other hand, this likely suggests a high information flow between visual tokens during the encoding process and a low token-level uniqueness or expressive intensity, such that spatially or randomly sampling tokens allows for capturing image-wide semantic concepts and relations. \\
Although we report results for 64 retained tokens in Fig. \ref{ElevenPercentResults} for LLaVA 1.5-7B, we find that these surprising trends persist with different visual retention rates and for a larger model (LLaVA 1.5-13B) as indicated by Tables \ref{table_llava7} and \ref{table_llava13}, although for the smaller LLaVA 1.5-7B model, the superior performance of ``Cluster \& Aggregate" is more pronounced for lower retention rates. In addition, Table \ref{table_vila} presents results from the VILA family of VLMs [11], which include only importance-agnostic methods operating at a fixed $11\%$ retention rate. Consistent with observations on the LLaVA models, our proposed ``Cluster \& Aggregate" compression variants achieve superior performance compared to these baselines.

    \begin{table}[ht]
\centering
\scriptsize
\caption{Performance of several visual token selection methods on various benchmarks for LLaVA1.5-7B}
\label{table_llava7}
\resizebox{1.0\columnwidth}{!}{ 
\begin{tabular}{@{}l c c c c c c c@{}}
\toprule
\textbf{Method} & \textbf{SQA IMG} & \textbf{TextVQA} & \textbf{POPE} & \textbf{MME} & \textbf{MMBench} & \textbf{MMVet} & \textbf{GQA} \\ 
\midrule
\multicolumn{8}{l}{\textbf{Baseline}} \\
LLaVA1.5-7B & 69.46 & 58.19 & 85.9 & 1862 & 64.6 & 31.3 & 61.93 \\
\midrule
\multicolumn{8}{l}{\textbf{Retained visual token count = 128}} \\
Random Sampling & 68.17 & 50.68 & 80.92 & 1663.93 & 59.8 & 30 & 57.21 \\
\textbf{Cluster \& Aggregate} & \textbf{68.96} & 55.25 & \textbf{86.14} & 1729.42 & \textbf{62.8} & 30.9 & \textbf{58.76} \\
SparseVLM & 67.1 & 54.9 & 80.5 & 1696 & 60 & 30 & 56 \\
FastV & 60.2 & 50.6 & 59.6 & 1490 & 56.1 & 28.1 & 49.6 \\
VisionZip & 68.9 & \textbf{56.8} & 83.2 & \textbf{1761.7} & 62 & \textbf{32.6} & 57.6 \\
\midrule
\multicolumn{8}{l}{\textbf{Retained visual token count = 192}} \\
Random Sampling & 69.51 & 52.13 & 82.68 & 1704.12 & 61.51 & 30.6 & 58.19 \\
Cluster \& Aggregate & 68.82 & 55.7 & \textbf{87.09} & 1744.44 & \textbf{63.32} & 29.4 & 59.16 \\
SparseVLM & 69.1 & 56.1 & 83.6 & 1721 & 62.5 & 31.5 & 57.6 \\
FastV & 67.3 & 52.5 & 64.8 & 1612 & 61.2 & 27.7 & 52.7 \\
VisionZip & \textbf{68.9} & \textbf{57.3} & 85.3 & \textbf{1782.6} & 63 & \textbf{31.7} & \textbf{59.3} \\
\bottomrule
\end{tabular}
}
\end{table}

\begin{table}[H]
\scriptsize
\caption{Performance of several visual token selection methods on various benchmarks for LLaVA1.5-13B.}
\label{table_llava13}
\resizebox{1.1\columnwidth}{!}{ 
\begin{tabular}{@{}l c c c c c c c@{}}
\toprule
\textbf{Method} & \textbf{SQA IMG} & \textbf{TextVQA} & \textbf{POPE} & \textbf{MME} & \textbf{MMBench} & \textbf{MMVet} & \textbf{GQA} \\ 
\midrule
\multicolumn{8}{l}{\textbf{Baseline}} \\
LLaVA1.5-13B & 72.68 & 61.25 & 86.04 & 1823.59 & 68.47 & 36.7 & 63.26 \\
\midrule
\multicolumn{8}{l}{\textbf{Retained visual token count = 64}} \\
Random Sampling & 71.34 & 50.78 & 72.68 & 1730.73 & 57.9 & 25.7 & 55.9 \\
Cluster \& Aggregate & 72.04 & 55.05 & \textbf{81.51} & \textbf{1731.25} & \textbf{65.4} & 31.8 & \textbf{58.77} \\
\textbf{VisionZip} & \textbf{74.4} & \textbf{57.4} & 76 & 1676 & 64.9 & \textbf{33.9} & 56.2 \\
\midrule
\multicolumn{8}{l}{\textbf{Retained visual token count = 128}} \\
Random Sampling & 72.24 & 52.73 & 78.79 & 1693.86 & 62.88 & 29.8 & 57.58 \\
Cluster \& Aggregate & 72.98 & 56.84 & \textbf{85.61} & \textbf{1771.36} & \textbf{66.75} & 33.4 & \textbf{58.94} \\
\textbf{VisionZip} & \textbf{74} & \textbf{58.7} & 85.2 & 1743 & 66.7 & \textbf{37.5} & 57.9 \\
\midrule
\multicolumn{8}{l}{\textbf{Retained visual token count = 192}} \\
Random Sampling & 72.58 & 54.22 & 81.55 & 1733.31 & 64.69 & 32.8 & 58.62 \\
\textbf{Cluster \& Aggregate} & \textbf{73.77} & \textbf{57.53} & 85.76 & \textbf{1810.62} & \textbf{67.35} & 33.8 & 59 \\
VisionZip & 73.5 & 59.5 & \textbf{85.9} & 1754 & 66.9 & \textbf{37.5} & \textbf{59.1} \\
\bottomrule
\end{tabular}
}
\end{table}

\begin{table}[H]
\centering
\scriptsize
\caption{Performance of several visual token selection methods on various benchmarks for VILA-8B.}
\label{table_vila}
\resizebox{1.0\columnwidth}{!}{ 
\begin{tabular}{@{}l c c c c c c c c@{}}
\toprule
\textbf{Method} & \textbf{SQA IMG} & \textbf{TextVQA} & \textbf{POPE} & \textbf{Vizwiz} & \textbf{MME} & \textbf{MMBench} & \textbf{MMVet} & \textbf{GQA} \\ 
\midrule
\multicolumn{8}{l}{\textbf{Baseline}} \\
VILA-8B & 81.95 & 68.54 & 85.69 & 63.24 & 1744.41 & 66.83 & 40.5 & 30.41\\
\midrule
\multicolumn{8}{l}{\textbf{Retained visual percentage 11\% }} \\
Random Sampling & 76.6 & 50.83 & 67.24 & 60.67 & 1448.93 & 48.45 & 26.7 & 41.53 \\
Spatial Sampling & 74.86 & 49.94 & 67.21 & 59.7 & 1423.29 & 50.1 & 27.4 & 36.14\\
Cluster \& Aggregate & \textbf{77.94} & \textbf{56.38} & \textbf{85.69} & \textbf{64.14} & \textbf{1566.2} & \textbf{55.84} & \textbf{32.2} & \textbf{44.02}
 \\

\bottomrule
\end{tabular}
}
\end{table}

\section{Conclusion}
Our initial experiments on visual token sequence compression highlight the need for efficient multimodal encoding due to the redundancy induced by the visual encoding pipeline and the computational costs of long context lengths. In particular, we have found that simple cluster-based token aggregation outperforms all previous finetuning-free SoTA methods for token selection and merging while being computationally simpler, and that spatial and random sampling perform better than most. Our observations also suggest that attention-based token importance is volatile and misaligned with intuition. This calls to revisit the growing intuition and reliance on attention scores for token selection.
We also found that certain visual tokens inherently attract more attention and therefore always appear more salient despite variations in the text prompt.
Closing the accuracy gap with baselines requires continued characterization of cross-modal token correlations and compression beyond token selection and clustering. We also aim to quantify the computational overhead and net gains.
We hope our findings will help orient new research in visual data encoding and compression for more efficient multimodal computing.\\ \\

\textbf{Acknowledgments}
We thank Tsachy Weissman, Mert Pelanci, and Pavlo Molchanov for their insightful suggestions and helpful feedback.\\ \\


{
    \small
    \bibliographystyle{ieeenat_fullname}
    \bibliography{multimodal}
}


\end{document}